\documentclass[wcp]{jmlr}

\usepackage{longtable}% for long tables
\usepackage{multirow}
\usepackage{booktabs}
\usepackage{xcolor}
\usepackage{arydshln}
\makeatletter
\let\Ginclude@graphics\@org@Ginclude@graphics 
\makeatother

\jmlrvolume{189}
\jmlryear{2022}
\jmlrworkshop{ACML 2022}

\title[Cross-Scale Context Extracted Hashing for Fine-Grained
Image Binary Encoding]{Cross-Scale Context Extracted Hashing for Fine-Grained
Image Binary Encoding}

\author{\Name{Xuetong Xue}  \footnotemark[1]  \Email{xuexuetong@corp.netease.com}\\
  \Name{Jiaying Shi} \footnotemark[1] \Email{shijiaying@corp.netease.com}\\
  \Name{Xinxue He} \Email{hexinxue@corp.netease.com}\\
  \Name{Shenghui Xu} \footnotemark[2] \Email{xushenghui@corp.netease.com}\\
  \Name{Zhaoming Pan} \Email{panzhaoming@corp.netease.com}\\
    \addr No.7 Building, Zhongguancun Software Park West, No.10 Xibeiwang East RD, Beijing, China}

\editors{Emtiyaz Khan and Mehmet G\"{o}nen}
\begin{document}

\maketitle

\renewcommand{\thefootnote}{\fnsymbol{footnote}} 
\footnotetext[1]{These authors contributed equally to this work.} %[1]
\footnotetext[2]{Corresponding author.} %[2]
\footnotetext[4]{Code: https://github.com/NetEase-Media/CSCE-Net}
%[3]

\begin{abstract}
Deep hashing has been widely applied to large-scale image retrieval tasks owing to efficient computation and low storage cost by encoding high-dimensional image data into binary codes. Since binary codes do not contain as much information as float features, the essence of binary encoding is preserving the main context to guarantee retrieval quality. However, the existing hashing methods have great limitations on suppressing redundant background information and accurately encoding from Euclidean space to Hamming space by a simple sign function. In order to solve these problems, a Cross-Scale Context Extracted Hashing Network (CSCE-Net) is proposed in this paper. Firstly, we design a two-branch framework to capture fine-grained local information while maintaining high-level global semantic information. Besides, Attention guided Information Extraction module (AIE) is introduced between two branches, which suppresses areas of low context information cooperated with global sliding windows. Unlike previous methods, our CSCE-Net learns a content-related Dynamic Sign Function (DSF) to replace the original simple sign function. Therefore, the proposed CSCE-Net is context-sensitive and able to perform well on accurate image binary encoding. We further demonstrate that our CSCE-Net is superior to the existing hashing methods, which improves retrieval performance on standard benchmarks.
\end{abstract}

\section{Introduction}

Hash encoding refers to compressing high-dimensional image pixels or feature points into binary codes instead of continuous features ~\cite{wang2015learning}. Due to its fast retrieval speed and low storage cost, deep hashing methods \cite{lai2015simultaneous, 2015Supervised, li2015feature, liu2016deep, xia2014supervised} have attracted lots of attention and been applied in the large-scale image and video retrieval to search similar ones from millions, which is quite common in today’s world \cite{liong2016deep}. Recently, with the help of CNN’s powerful representation capability, deep hashing methods \cite{yuan2020central,yang2021dolg} have shown excellent performance on benchmarks.
The existing methods often learn to utilize pairwise or triplet data similarity to encode images.
In order to retain the discrimination of the original data, the concepts of ``Hash Center'' \cite{yuan2020central} and ``Proxy'' \cite{wieczorek2021unreasonable} are proposed,  which can reduce the computational complexity and simplify the problem to optimize the intra-class distance only by finding a set of orthogonal vectors as hash centers.

The most existing deep hashing algorithm \cite{liu2016deep, hoe2021one, yuan2020central} firstly obtains continuous float features through deep network's (eg. Alexnet \cite{krizhevsky2012imagenet} or ResNet50 \cite{he2016resnet}) last fully connect layer before classification layer, then computes binary codes (0, 1) by a simple sign function as post-processing. Therefore, network can be optimized by pairwise or triplet based loss \cite{cao2017hashnet, zhu2016deep, li2015feature} as classification tasks or cosine similarity loss as metric learning tasks. With the emergence of deeper and more complex networks, such as Transformer \cite{vaswani2017attention}, deep learning begins to pay attention to finding stronger feature representations. For example, \cite{balntas2016learning, noh2017large, revaud2019r2d2, yang2021dolg} introduced discriminated local features to guide the network to classify difficult samples through fine local information. %Besides, numerous researchers have carried out in-depth efforts to tackle information\cite{hoe2021one}. 
%(As shown in figure 1, the high-level semantic global representation learned by deep networks is prone to recall error samples with insignificant targets or complex backgrounds. With the emergence of deeper and more complex networks, such as Transformer \cite{vaswani2017attention}, deep learning began to pay attention to finding stronger feature representations. \cite{balntas2016learning, noh2017large, revaud2019r2d2, yang2021dolg} proposed to introduce discriminated local features to guide the network to classify difficult samples through fine local information, and at the same time in order to reduce information redundancy, \cite{yang2021dolg} adopted orthogonal fusion to improve the robustness of features.)
\begin{figure}[htbp]
\centering
\includegraphics[width=0.8\textwidth]{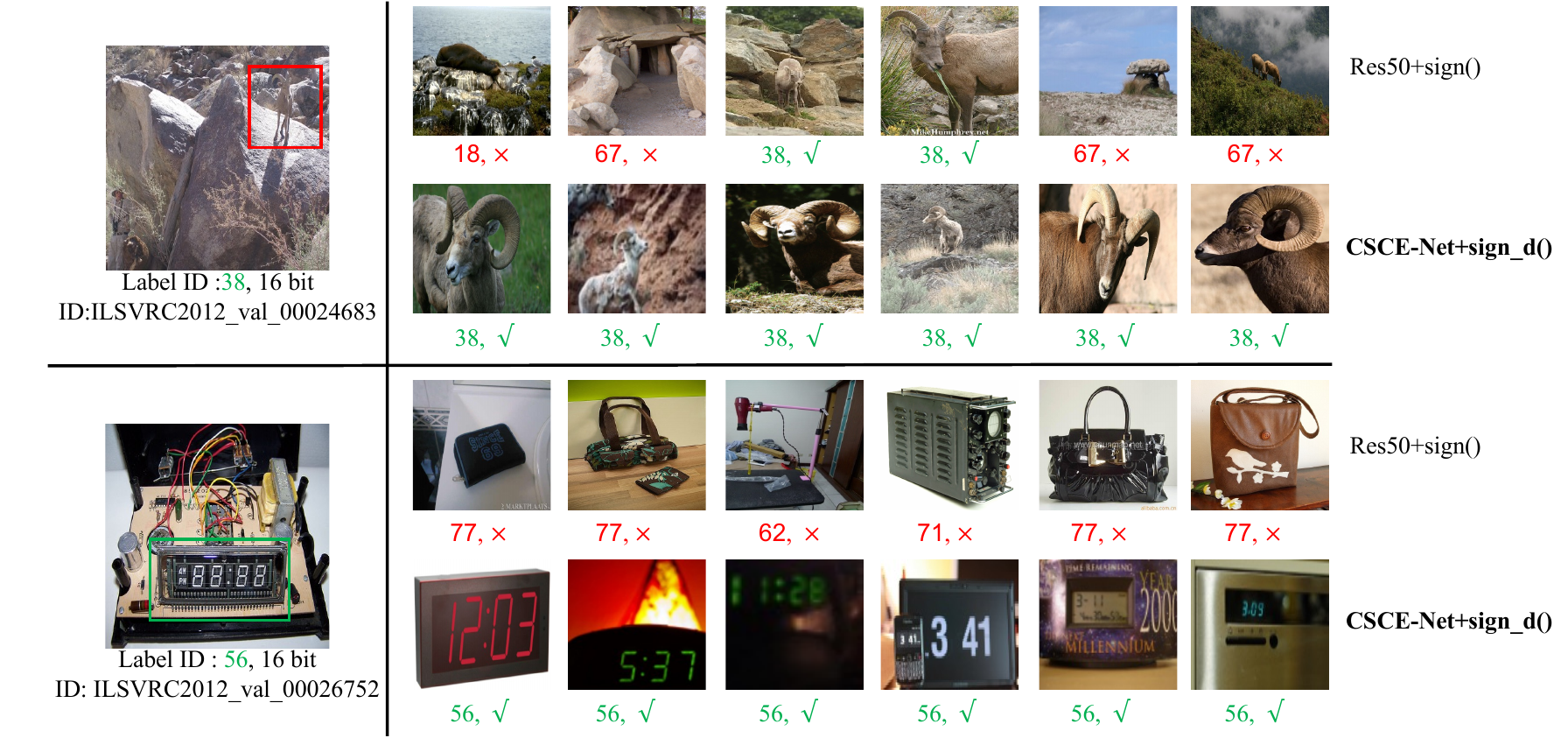}
\caption{Visualization of two cases for $standard$ hashing method and proposed CSCE-Net. Simple $CNN+sign()$ methods encode entire image with no difference resulting in error recalls with similar background, for example label 18, 67 in case one. In contrast, the proposed CSCE-Net can encode key subject rather than background accurately.}
\label{fig:fig1}
\end{figure}

Although these approaches attempted to make network learning a more compact representation, they ignored the relatively weak expressiveness of binary code. Hence, one challenge that needs to be addressed is how to maintain the most important information in binary codes from original float features. Besides, Fig.~\ref{fig:fig1} shows two example cases, where the pure high-level semantic global representation learned by deep network is prone to recall error samples with unapparent targets (red box) or apparent targets (green box) with complex backgrounds. That is to say, network does not give enough representation for important targets when encoding images. Therefore, another challenge is how to make the network automatically extract the most critical semantic context contained in images to produce better representations.

In this paper, we propose a Cross-Scale Context Extracted hashing Network (CSCE-Net) to extract the relative significant context of targets for image retrieval, which consists of two branches, aiming to jointly align global semantics and local details. Cooperated with sliding windows, an Attention guided Information Extraction module (AIE) is introduced between two branches to suppress low information areas by utilizing multi-scale semantic global context to filter fine-grained local information. Besides, instead of applying a simple sign function, CSCE-Net learns a content-related Dynamic Sign Function (DSF) to further ensure the accuracy of binary encoding. Generally speaking, our main contributions are as follows:

\begin{enumerate}

    \item[$\bullet$] We first consider the representation capability of binary code and propose CSCE-Net to automatically extract critical image information. 
    \item[$\bullet$] An Attention guided Information Extraction module (AIE) is designed to further utilize global semantics to extract fine-grained local information, which improves the quality of handling unapparent targets and complex backgrounds.
    \item[$\bullet$] A content-related Dynamic Sign Function module (DSF) is designed to encode continuous features into binary codes, which reduces information loss adaptively.
    %In order to adaptively maintain stability of converting float feature to binary code, a content-related Dynamic Sign Function (DSF) is designed in our CSCE-Net.
    % We introduce a content-related Dynamic Sign Function (DSF) which makes hash code adaptive for representing hash code in 
    % \textcolor{red}{
    % Experiments show that the proposed MSISNet has better performance than existing salient instance segmentation methods.
    \item[$\bullet$]  We conduct extensive experiments to demonstrate that the proposed CSCE-Net outperforms existing methods and achieves remarkable state-of-the-art performance on three public benchmarks.
    % \item[$\bullet$]  Experiments show that the proposed CSCE-Net has better performance than existing n three standard datasets including ImageNet100, NUS-WIDE, and MS COCO by x% - x%, x% - x% and x%-x% in MAP, respectively.
    % }
\end{enumerate}

\section{Related Work}

The deep hashing methods have been activately researched, such as CNNH \cite{xia2014supervised}, DNNH \cite{lai2015simultaneous}, DHN \cite{zhu2016deep}, HashNet \cite{cao2017hashnet}, DCH \cite{cao2018deep}, CSQ \cite{yuan2020central}, DOLG \cite{yang2021dolg}. 
Among these related works, the early methods usually adopt pairwise or triplet similarity learning to encode images. CNNH \cite{xia2014supervised}, a two-stage method, decomposed the sample similarity matrix to obtain the binary code of samples, then used CNN to fit the binary codes. 
%Since it is not end-to-end training, this method based on manual design features does not take full advantage of deep learning despite its improved performance.
Furthermore, the deep architecture proposed by DNNH \cite{lai2015simultaneous} used a triplet ranking loss to preserve relative similarities which converted the input images into unified image representation and then encoded them into hash codes by divide-and-encode modules.
DHN \cite{zhu2016deep} simultaneously optimized the pairwise cross-entropy loss on semantic similarity pairs and the pairwise quantization loss on compact hash codes. HashNet \cite{cao2017hashnet} generated binary hash codes by optimizing a novel weighted pairwise cross-entropy loss function in deep convolutional neural networks.
However, due to the pair-wise data limitation, these pair-wise or triplet methods have insufficient coverage for data distribution and low efficiency across the entire training dataset.
%DCH \cite{cao2018deep} generated compact and concentrated hash codes by jointly optimizing a novel Cauchy cross-entropy loss and a Cauchy quantization loss in a single Bayesian learning framework.

To address the above limitations, CSQ \cite{yuan2020central} proposed a novel concept “Hash Center” and formulated the central similarity for deep hash learning which transformed hash code similarity learning to classification tasks.
\cite{hoe2021one} unified training objectives of deep hashing under a single classification objective by maximizing the cosine similarity between the continuous codes and binary orthogonal target under a cross-entropy loss. 
Besides, with the same basic idea of maximizing feature similarity,  CosFace \cite{wang2018cosface} and Arcface \cite{deng2019arcface} are also mentioned in hashing tasks.
% DONE (\textbf{TODO}, mention face rec work here, cite cosface, arcface, centerface etc.). 
The concept of above methods is to learn the high-level semantic features of images through deep networks, and then use a hash network to generate compact binary codes. However, for similar samples with much redundant background, the global representation obtained by deep learning cannot be accurately retrieved, requiring more refined local features for recognition. 

Therefore, in recent years, several methods focus on how to represent an image with much richer information. \cite{simeoni2019local,noh2017large,cao2020unifying} leveraged global features to select candidate images and then performed fine matching on candidate images according to local features.
DELF \cite{noh2017large} designed an attentive local feature descriptor and an attention mechanism for key point selection to identify usefully semantic local features for image retrieval.
DELG \cite{cao2020unifying} proposed a unified model which leveraged generalized mean pooling to produce global features and attention-based key point detection to produce local features.
%Although local features make it easier to retrieve difficult samples, these two-stage algorithms may lead to error accumulation, and the algorithms used to extract local features, such as RANSAC, is time-consuming.
DOLG \cite{yang2021dolg} attempted to fuse local and global features in an orthogonal manner for effective single-stage image retrieval. 
%DSaH \cite{jin2018deep} proposed an attention network which could automatically mine salient regions and learn semantic-preserving hashing codes simultaneously.

Inspired by above concepts, the proposed method designs a two-branch feature extraction architecture among different scales. Interacted by an Attention guided Information Extraction module (AIE), our CSCE-Net aligns local and global spatial context to give enough representation for important targets when encoding images, which is contrary to DOLG \cite{yang2021dolg} focusing on feature fusion. Despite this, all the above methods use a sign function to encode features from Euclidean space to Hamming space, which leads to the quantization loss of binary code. Therefore, we consider the compactness between float and binary codes with a content-related Dynamic Sign Function to extend the ability from float features.

% DONE BY SJY (\textbf{TODO},mention difference between CSEF-Net and DOLG:1.different ways of interaction between two branch, dolg fusion feature,CSEF-NEt extract  context information;2 CSEF-Net keep global spatial information with align local  spatial information, for cross-scale guide information extract)

\section{Method}

In this paper, we propose a Cross-Scale Context Extracted Hashing Network (CSCE-Net), which is consist of an Attention guided Information Extraction module (AIE) and a content-related Dynamic Sign Function (DSF). The proposed framework is described in detail as follows:

\subsection{Problem Definition}
As mentioned above, deep hash networks usually learn a compact continuous feature and then transform it into hash code by sign function. 
We define the deep hashing problems of $N$ training samples and C categories as follows:
we express
the original $D$-dimensional image data as 
$X = \{x_{i}\}_{i=1}^{N} \in \mathbb{R}^{N \times D}$ 
and denote the one-hot training corresponding semantic labels as $Y = \{y_{i}\}_{i=1}^{N} \in \{0,1\}^{N \times C}$.
For each training sample $(x_{i},y_{i})$, we express the compact continuous features generated by the deep network as $f_{e_{i}}$, and represent the $K$-bit binary codes $b_{i} \in \{-1,1\}^{N \times K}$ through sign function as shown in Fig.~\ref{fig:pipeline}.
Following previous work in CSQ \cite{yuan2020central} and ICS \cite{zhang2021instance}, our model also adopts the hash center
$\mathcal{H}=\{h_{i}\}_{i=1}^{C} \in\{-1,1\}^{C \times K}$ to convert $C$ class labels to the corresponding Hadamard matrix rows with $K$-bit, which guarantees the orthogonality of any two class centers and facilitates high precision retrieval as previous work has proven.

% Following \cite{yuan2020central} and \cite{zhang2021instance}, we define the row vectors in Hadamard matrix $H_{k}$ as hash centers. Due to the nature of $H_{k}$ satisfying row vectors and column vectors are pair-wise orthogonal, and half of bits are 1 or -1. We can simply replace all -1 with 0 to obtain hash centers in {0, 1}

The hash center can be generated by 2k-order Hadamard matrix $H$  simply using the Sylvester’s algorithm \cite{sylvester1867lx}: 
\begin{equation}\label{eq-hardmard}
H{_{2}}^{k} = \begin{bmatrix}
 H{_{2}}^{k-1} &  H{_{2}}^{k-1}\\ 
 H{_{2}}^{k-1} & -H{_{2}}^{k-1} 
\end{bmatrix} =H_{2} \bigotimes H{_{2}}^{k-1},   \\
H_{2} = \begin{bmatrix}
1 & 1\\ 
1 & -1
\end{bmatrix},
\end{equation}

where $\bigotimes$ represents the Kronecker product, $2^k$ is the number of orthogonal centers.

\subsection{Overview of CSCE-Net}

\begin{figure}[htbp]
\centering
\includegraphics[width=0.7\textwidth]{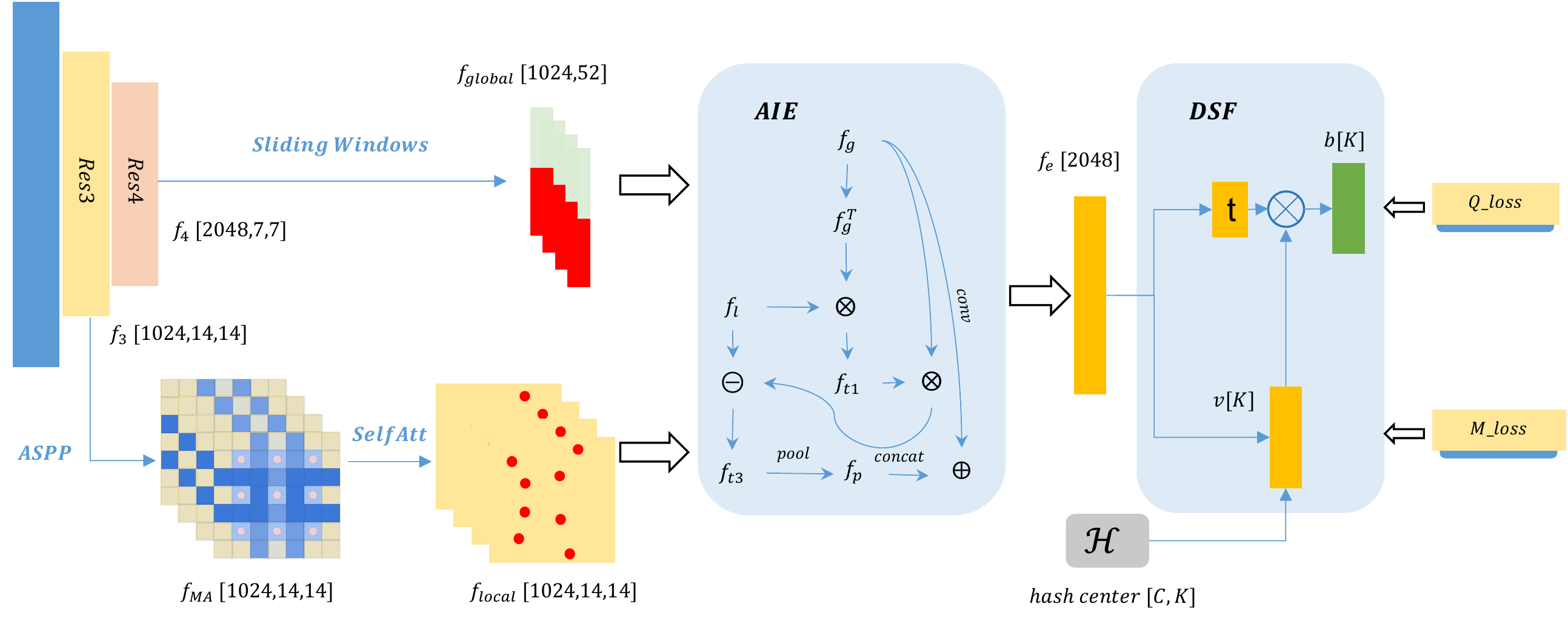}
\caption{Overall structure of proposed CSCE-Net, built upon ResNet50. The global branch takes the output of $Res4$ block as input to get multi-scale features $f_{global}$. The local branch uses ASPP and SelfAtt to model more local details $f_{local}$ after $Res3$ block. AIE takes both as input to generate final representation $f_e$. $Q\_loss$ and $M\_loss$ are quantization loss and metric loss respectively.}
\label{fig:pipeline}
\end{figure}

In order to solve the problem that the previous network does not give enough 
representation to unapparent targets when coding images, a two-branch framework 
CSCE-Net is designed. Following \cite{zhu2016deep, cao2017hashnet, yuan2020central}, it is built upon state-of-the-art image recognition model ResNet50.
As shown in Fig.~\ref{fig:pipeline}: 1) The global branch keep the same as the original ResNet50 except that we remove all $pooling$ and $fc$ layers after $Res4$ block $f_{4} \in R^{\mathfrak{c}_{4} \times h \times w}$ to keep spatial information in feature maps, and the multi-scale sliding windows are introduced to get multi-scale spatial global information; 2) The major building blocks of our local branch (start from $Res3$ block $f_{3} \in R^{\mathfrak{c}_{3} \times h \times w}$) is based on ASPP \cite{chen2017aspp} and self-attention mechanism \cite{noh2017selfatt}. Then AIE module is used to further model the semantics of global window features while helping to extract fine-grained local feature. The global features $f_{global}$ abbreviated as $f_g$ can be formulated as:

\begin{equation}\label{eq-infty}
    f_{g}=concatenate(torch.unfold(f_{4_1},sw1), torch.unfold(f_{4_2},sw2)),
\end{equation}

\begin{figure}[htbp]
\centering
\includegraphics[width=0.4\textwidth]{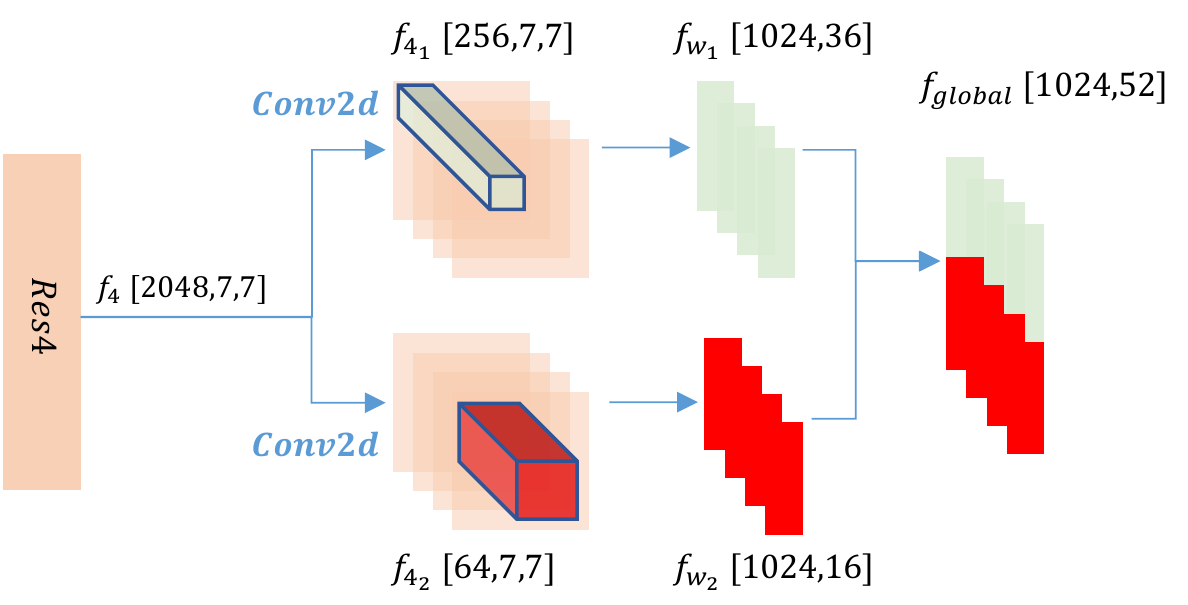}
\caption{The structure of sliding window in global branch based on $Res4$ block with two different window sizes.}
\label{fig:slidingwindow}
\end{figure}

where $f_{4_1}$ and $f_{4_2}$ have different feature size after $Conv2d$ as shown in Fig.~\ref{fig:slidingwindow}, and $sw1$ and $sw2$ are the hyper parameters of window size, which we use $2\times2$ and $4\times4$ respectively.
Specifically, for the local branch, a multi-atrous dilated convolution layer is introduced in ASPP to handle scale variations among different image instances, which outputs feature maps with the different spatial receptive fields. Then a self-attention module can model the importance of each feature point in spatial. As is known to all, the shallow layers of the deep network focus on detailed information such as edges and corners, while the high layers perceive more overall and semantic information. Thus, the intuition behind two-branch design is to take advantage of the semantic context of $Res4$ to align and filter better local features to suppress interference of noise.
In terms of local structure,  ASPP module contains three dilated convolution layers to obtain $f_{MA}$ with different spatial receptive fields, then we compute $f_{local}$ with an attention map produced by a $1\times 1$ convolution layer and the SoftPlus operation. Each feature size is shown in Fig.~\ref{fig:pipeline}. Additionally, the proposed CSCE-Net keeps spatial information in both branches to align context adaptively which would be mentioned next.

\subsection{Attention based Information Extraction Module, AIE}

\begin{figure}[htbp]
\centering
\includegraphics[width=0.6\textwidth]{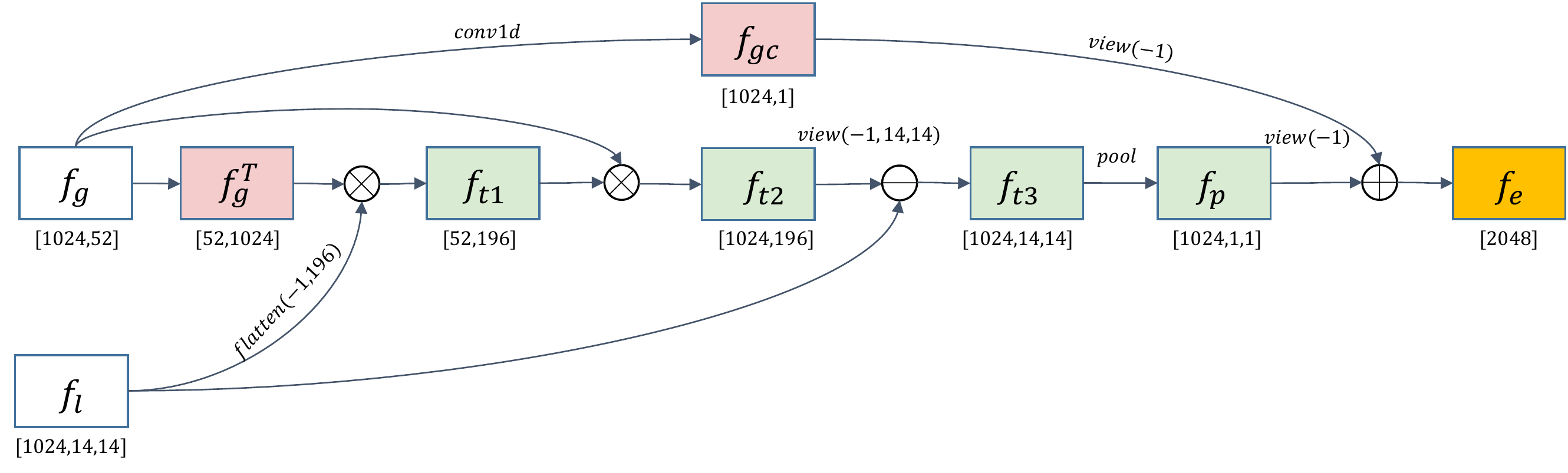}
\caption{Detailed structure of Attention guided Information Extraction module, AIE}
\label{fig:aie}
\end{figure}

Existing hashing methods often process features on the global branch, by $pooling$ or $fc$ layer,
which results in information loss on spatial. As shown in Fig.~\ref{fig:fig1}, simple $CNN+sign()$ lacks the ability to separate redundant context from images, especially for unapparent targets. Therefore, binary code inevitably encodes a lot of unnecessary information. In order to effectively capture important areas in spatial, an Attention guided Information Extraction module (AIE) follows after multi-atrous ASPP and $SelfAtt$.  The structure of AIE is shown in Fig.~\ref{fig:aie}.

The inputs of AIE include two parts: sliding windows based multi-scale global features $f_{global}$ abbreviated as $f_{g}$ and attention based multi-atrous local features $f_{local}$ abbreviated as $f_l$. First, through dot product operations ($torch.mm$), we align two features and compute the repetitive context from global to local which can be formulated as: $ f_{t2} = f_{g} \otimes  (f_{g}^{T} \otimes  f_{l} )$. Because of the fact that global features have more semantic context as well as local features have more detailed context, for unapparent targets in complex background, the repetitive context would contain much more useless semantics. Therefore, it is necessary to separate them from local details, so
secondly, we remove redundant features using $minus$ to extract fine-grained local feature with multi-scale spatial context from global information. The whole process can be formulated as:
%we conduct two dot product operations ($torch.mm$) among $f_{l}$ , $f_{g}$ and flatten $f_{l}$ to model ``correlation" information between two branch:
%The inputs of AIE include two parts: sliding windows based multi-scale global featues $f_{g}$ and attention based multi-atrous local features $f_{l}$. First, we conduct two dot product operation ($torch.mm$) between \textcolor{red}{$f_{l}$ , $f_{g}$ and flatten $f_{l}$, }to modeling "correlation" information between two branch:
\begin{equation}\label{e1}
f_{t3}  = f_{l} - f_{g} \otimes  (f_{g}^{T} \otimes  f_{l}).
\end{equation}
%\textcolor{red}{SHAPE}
%here, each feature shape is shown in Fig. \ref{fig:aie} 
%$f_{e}$ shape: $f_{l}$ shape:,  $f_{g}$ shape:

Then, we obtain extracted ``critical" information $f_{t3}$ from correlation modeling, and the final extraction feature $f_{e}$ is concated with global branch:

\begin{equation}\label{e2}
f_{e} = concat (conv1d(f_{g}).view(-1), pool(f_{t3}).view(-1)).
\end{equation}

In this way, the final extracted $f_{e}$ aligns the detailed local features with the global spatial context to focus on critical instance areas.

% DONE DELETE\textcolor{red}{todo:add description about pool conv ops to deal with dimention}

%(AIE enhances/provide ..., In addition, AIE adopts.... so as to alleviate the problem that.... Specifically(structure. In the last layer, xxxx. Finally, skip connection)

\subsection{Content-related Dynamic Sign Function, DSF}

As shown in Formula \ref{eq-sign}, the common binarization method sign function jumps around 0, which is equivalent to having a fixed threshold value of 0. The threshold cannot be changed according to the image content.

\begin{equation}\label{eq-sign}
    sign(x)=\left\{\begin{array}{l}
    1, x>0\\
    -1, else
    \end{array}\right.,
\end{equation}
where we use 1 and -1 to represent binary code which also can be 1 and 0. Therefore, we propose a hypothesis that if the threshold can be changed according to the content of the image, images which have similar float features in original space will have a higher chance of being closer in Hamming space. Therefore, we propose content-related Dynamic Sign Function namely DSF to learn a dynamic threshold for $sign()$. 

Although it is better for hash code to have a balanced bit of being $+1$ or $-1$ \cite{hoe2021one}, there is no guarantee that encoded hash code through $sign$ function also maintains this balance.  %Hadamard matrix makes sense in code balancing every bit has 50\% of chance being $+1$ or $−1$, there is no guarantee that a simple change in the threshold also maintains this balance. 
Therefore, we propose a learnable threshold $t (t\geq 0)$ , and dynamically determine the direction of the jump to -1 or 1. The threshold $t$ is proposed by a $fc$ layer with output size of 1 shown in ``DSF'' in Fig.~\ref{fig:pipeline}. For more details, according to learned $t$, we set an interval $[-t,t]$, then count the number of $1$ and -$1$ outside the interval to dynamically encode the value inside $[-t,t]$ to -1 or 1 for code balance.
%$dynamic$ is the non-dominant side. 
This rule ensures that $sign()$ function maintains balance as much as possible. The definition of the new dynamic $sign$ function is as follows:
\begin{equation}\label{eq-newsign}
    sign\_d(x)=\left\{\begin{array}{l}
    1, x>t\\
    dynamic, -t \leq x \leq t\\
    -1, x<-t
    \end{array}\right.,
\end{equation}
\begin{equation}\label{eq-dynamic}
    dynamic=\left\{\begin{matrix}
-1 ,& \frac{count(x=1)}{count(x=-1)} > 1\\ 
1 ,& else
\end{matrix}\right.,
\end{equation}
when the number of values bigger than $t$ is more than the number of values smaller than $-t$, all inside interval $[-t,t]$ will jump to -1, and vice versa. As shown in Fig.~\ref{fig:pipeline}, the threshold $t$ is learnt from extracted hash feature, which is content-related and context-sensitive. Note that for more stable training, the upper bound of $t$ was set to 0.005 as a hyper-parameter.
%\textcolor{red}{mention t de shangjie is a hyper params??}
\subsection{Loss Functions for Image Hashing}
\label{section_loss}

%\textcolor{red}{loss meiyougai, todo}
\textbf{Metric Loss.} 
Generally speaking, for any two $K$-dimension binary codes $b_{1}$, $b_{2}$, the Hamming distance estimates the similarity of the two binary codes by XOR operation per bit. As we all know that XOR is non-differentiable, which cannot be used for training networks by back propagation algorithm \cite{rumelhart1986learning}. Therefore, cosine similarity is adopted instead to approximate the distance between continuous float code $v_{1}$, $v_{2}$ for calculating the loss. Besides, the binary code can be computed by $sign$ function from continuous code, which is expressed in the formula as:

\begin{equation}\label{eq-sign2}
b_{i}=sign\_d(v_{i}),
\end{equation}

where $sign\_d$ function can be formulated as \ref{eq-newsign}. Therefore, the relationship between cosine similarity and Hamming distance is formulated as \cite{xu2021hhf}:

\begin{equation}\label{eq-cosin}
\mathfrak{D}_{ham} (b_{i}, b_{j}) = \frac{K}{2}(1 - \frac{b_{i}\cdot b_{j} }{\left \| b_{i}\right \|\cdot \left \|  b_{j} \right \|})\approx - \cos (v_{i}, v_{j}),
\end{equation}

where $K$ is the dimension of hash code. As is known, metric loss aims at clustering features by pushing different category samples as far as possible and pulling samples of the same category as close as possible. Following \cite{hoe2021one}, \cite{deng2019arcface},\cite{wang2018cosface}, the training of our CSCE-Net is also based on hash center and angular margin penalty-based cosine softmax loss (metric loss).  The continuous code $v_{i}$ also called logits can be reformulated as: $v_{i}$ = $f_{e_{i}}\cdot h_{i}^{T}=\left \| f_{e_{i}}\right \|\left \| h_{i}\right \|\cos (\theta _{h_{i}})$, where $\theta _{h_{i}}$ is the angle between the hash center $h_{i}$ and the extracted hash feature $f_{e_{i}}$ as shown in ``DSF" part of Fig.~\ref{fig:pipeline}. 
By the norm operation of extracted hash feature and hash center, we can get two equations, $\left \| f_{e_{i}}\right \|=1$ and $\left \| h_{i}\right \|=1$. Therefore, it is explained that we can set $\left \| v_{i}\right \|$ to a constant parameter $S$. 
The softmax cross-entropy loss is formulated as:

\begin{equation}\label{eq-softmax-ori}
    L_{ce} = \frac{1}{N}\sum_{i\in N}^{}-\log \frac{e^{v_{i}}}{\sum_{j=1}^{C}e^{v_{j}}},
\end{equation}
where $C$ is the category number. Taking the above formula of ${v_{i}}$  into account and setting a constant bias, we can get a transformed formula for $L_{ce}$,

\begin{equation}\label{eq-softmax}
    L_{ce} = \frac{1}{N}\sum_{i\in N}^{}-\log \frac{e^{f_{e_{i}}\cdot h_{i}^{T} + bias}}{\sum_{j=1}^{C}e^{f_{e_{j}}\cdot h_{j}^{T} + bias}} \\
    = \frac{1}{N}\sum_{i\in N}^{}-\log \frac{e^{S\cdot \cos(\theta_{h_{i}} )}}{e^{S\cdot (\cos(\theta_{h_{i} )})} + \sum_{j=1,j\neq i}^{C} e^{S\cdot\cos(\theta_{h_{j}})}},
\end{equation}

For simplicity, we fix bias offset equal to 0. $\theta_{h_{j}}$ is the angle between extracted hash feature $f_{e_{i}}$ and the other hash center $h_{j}$ for negative computation. 
Therefore, the general angular margin penalty-based metric loss $L$ is defined as follows:

\begin{equation}\label{eq-cos}
L = \frac{1}{N}\sum_{i\in N}^{}-\log \frac{e^{S\cdot (\cos(m_{1}\cdot \theta_{h_{i}} +m_{2})-m_{3})}}{e^{S\cdot (\cos(m_{1}\cdot \theta_{h_{i} }+ m_{2})-m_{3})} + \sum_{j=1,j\neq i}^{C} e^{S\cdot\cos(\theta_{h_{j}})}}.
\end{equation}

The angular margin penalty is different in SphereFace \cite{liu2017sphereface} ($m_{1} = \alpha , m_{2} = 0 , m_{3} = 0 , \alpha > 1.0 $), 
CosFace \cite{wang2018cosface} ($m_{1} = 1 $, $m_{2} = 0 $, $m_{3} = \alpha, 0 < \alpha < 1 - \cos (\frac{\pi}{4})$) and Arcface \cite{deng2019arcface} ($m_{1} = 1 $, $m_{2} = \alpha $, $m_{3} = 0,  0< \alpha < 1.0 $). We use CosFace margin loss in this paper to train the whole CSCE-Net as formulated as:

\begin{equation}\label{eq-cos-mtric}
M\_loss = \frac{1}{N}\sum_{i\in N}^{}-\log \frac{e^{S\cdot (\cos(\theta_{h_{i}})-\alpha)}}{e^{S\cdot (\cos(\theta_{h_{i} })-\alpha)} + \sum_{j=1,j\neq i}^{C} e^{S\cdot\cos(\theta_{h_{j}})}}.
\end{equation}

\textbf{Quantization Loss.}
Metric loss only guarantees the continuous codes have favorable intra-class compactness and inter-class separability in original space. In order to mitigate the loss of information caused by $sign()$ operation, quantization loss \cite{zhu2016deep} is proposed to constrain latent codes.
By adding quantization loss \cite{li2017deep} \cite{yuan2020central} \cite{zhe2019deep} \cite{wang2021weakly}, retrieval performance is demonstrated to achieve promising improvement. Quantization loss $Q\_loss$ can be formalized as:

\begin{equation}\label{l_q}
Q\_loss =  \frac{1}{N}\sum_{i=1}^{N}\left \| v_{i} - b_{i} \right \| _{2}^{2} =\frac{1}{N}\sum_{i=1}^{N}\left \| v_{i} - sign\_d(v_{i}) \right \| _{2}^{2} ,
\end{equation}
where $v_{i}$ is continuous codes and $b_{i}$ is corresponding binary codes formulated by Equation \ref{eq-sign2}.
%In Hamming space, benefiting from dynamic $sign\_d()$, some bits close to 0 of similar samples would not be ambiguously binary encoded to ±1, which is highly desirable for clustering images with similar hash codes. \textcolor{red}{TODO: confuse}.
By combining metric loss and quantified loss, our learning objective can be expressed as:
 \begin{equation}\label{eq-GenLoss}
     \min (M\_loss+\lambda Q\_loss),
\end{equation}

where we use $\lambda=1$ to train CSCE-Net in an end-to-end manner.

\section{Experiments}
In this section, the proposed CSCE-Net would be evaluated and compared with several state-of-the-art hashing methods. We firstly introduce three popular datasets we used, namely ImageNet100, MS COCO, and NUS-WIDE, and then describe the evaluation protocols, implementation details, experiment comparisons, and ablation studies respectively.

\subsection{Dataset}
To verify the performance of our proposed method, we compare our method with the state-of-the-art methods by conducting extensive experiments on three widely-used benchmarks following prior works \cite{cao2017hashnet, cao2018deep, yuan2020central}, which includes ImageNet100 \cite{russakovsky2015imagenet}, MS COCO \cite{lin2014microsoft}, and NUS-WIDE \cite{chua2009nus}. In terms of the hash center for classification, we generate each category center referring to CSQ \cite{yuan2020central} for both single-label and multi-label datasets.

\textbf{ImageNet100} is a large-scale benchmark for the visual recognition challenge, which contains over 1M images with a single label totally. For a fair comparison, we use the same data and settings as \cite{cao2017hashnet, yuan2020central}, containing 10K, 5K, 120K for training, testing and retrieval respectively in 100 categories.

\textbf{MS COCO} is a multi-label benchmark containing about 120K images belonging to 80 category types. Following \cite{cao2017hashnet}, after pruning
images with no category information, we randomly sample 10K, 5K, 110K for training, testing and retrieval respectively.

\textbf{NUS-WIDE} is a public Web image dataset consisting of 269,648 multi-label images. Actually, this dataset contains 81 ground truth concept labels and we selected 21 most frequent categories following \cite{zhu2016deep} to sample 10K, 2K, 140K for training, testing and retrieval respectively.

\subsection{Evaluation Protocols and Implementation Details}

We evaluate the retrieval performance based on the widely used metric mean average precision (mAP). For a fair comparison, we follow the prior methods \cite{zhu2016deep, yuan2020central} that adopt mAP@1000 for ImageNet and mAP@5000 for the other datasets. The mAP is able to measure retrieval quality reliably which provides an average result of recall performance across all samples.

Following previous works, all the experiments of the proposed method in this paper are trained based on ResNet50 initialized from ImageNet pretrained weights and we fine-tune convolutional layers with proposed AIE and DSF modules through back propagation. As the proposed modules are trained from scratch, we use RMSProp optimizer with $ 1e^{-5}$ initial learning rate and $ 1e^{-5}$ weight decay factor. For the CosFace margin loss, we empirically set the margin $\alpha$ as 0.15 and the CosFace scale $S$ as 10. The images are first resized to 256 × 256 resolution and then center cropped to 224. All experiments use batch size of 128 trained on a single A10 GPU with 24G memory for 100 epochs.

% \begin{table}[t]
% \centering

% \begin{tabular}{llp{5mm}<{\centering}p{5mm}<{\centering}p{5mm}<{\centering}p{5mm}<{\centering}p{5mm}<{\centering}p{5mm}<{\centering}}
% \hline
% \multicolumn{2}{c}{ \multirow{2}*{Location} }& \multicolumn{3}{c}{Roxf} &\multicolumn{3}{c}{Rpar}\\
% \cline{3-8} 
% \multicolumn{2}{c}{} &E &M & H &E &M &H  \\
% \toprule
% \multicolumn{2}{l}{Global only}&90.65 &78.21 &56.31 &95.65 &89.00 &76.17  \\
% \multicolumn{2}{l}{Fuse f4-only}&92.08 &79.39 &58.13 & 95.93&\textbf{89.92} & \textbf{77.92} \\ 
% \multicolumn{2}{l}{Fuse f3-only}&\textbf{93.17} &\textbf{80.50} &\textbf{58.82} &95.95 &89.81 &77.70  \\ 
% \multicolumn{2}{l}{both f3\&f4}&92.34 &79.41 &57.08 &\textbf{96.01} &89.78 &77.69  \\ 
% \hline
% \end{tabular}
% \caption{}
% \label{t-2-location-ablation}
% \end{table}

\subsection{Experiment Results}
\subsubsection{Comparison with State-of-the-art Methods}

\begin{table}[h]
\centering
\scriptsize
\caption{Comparison in mAP of Hamming Ranking for different bits on image retrieval. }
%The results  from the original paper are marked with ``\underline{      }".} 
% \setlength\tabcolsep{1pt}

\begin{tabular}{ccccc|ccc|ccc}
    \hline
    \multicolumn{1}{l}{\multirow{2}*{Method}} &  \multicolumn{1}{l|}{\multirow{2}*{Backbone}} & \multicolumn{3}{l|}{ImageNet (mAP@1000)} &\multicolumn{3}{c|}{MS COCO(mAP@5000)} & \multicolumn{3}{c}{NUS-WIDE (mAP@5000)}\\
    
    \cline{3-11} 
    \multicolumn{2}{l}{}{} & 16bits & 32bits & 64bits & 16bits & 32bits & 64bits & 16bits & 32bits & 64bits \\
    \toprule
        \multicolumn{1}{l}{TransHash}& ViT & 0.785 & 0.873 & 0.892 & - & - & - & 0.726 & 0.739 &
        0.749 \\ 
        \multicolumn{1}{l}{CIBHash } & VGG16 & 0.718 & 0.756 & 0.783 & 0.737 & 0.760 & 0.775 & 0.790 & 0.807 & 0.815  \\  
        % \multicolumn{1}{l}{CIBHash } & ViT & - & - & - & 0.809 & 0.846 & 0.867 & 0.779 & 0.810 & 0.826  \\  
        \multicolumn{1}{l}{CIBHash } & GCViT-XXT & 0.860 & 0.886 & 0.897 & 0.807 & 0.836 & 0.851 & 0.795 & 0.817  & 0.825  \\

        \hdashline
        \multicolumn{1}{l}{CNNH  }  & ResNet50 & 0.315 & 0.473 & 0.596 & 0.599 & 0.617 & 0.620 & 0.655 & 0.659 & 0.647 \\
        \multicolumn{1}{l}{DNNH }  & ResNet50 & 0.353 & 0.522 & 0.610 & 0.644 & 0.651 & 0.647 & 0.703 & 0.738 & 0.754 \\
        \multicolumn{1}{l}{DHN}  & ResNet50 & 0.367 & 0.522 & 0.627 & 0.719 & 0.731 & 0.745 & 0.712 & 0.759 & 0.771 \\
        \multicolumn{1}{l}{HashNet }  & ResNet50 & 0.622 & 0.701 & 0.739 & 0.745 & 0.773 & 0.788 & 0.757 & 0.775 & 0.790 \\
        \multicolumn{1}{l}{CSQ }  & ResNet50 & 0.851 & 0.865 & 0.873 & 0.796 & 0.838 & 0.861 & 0.810 & 0.825 & 0.839 \\

    \toprule
        \multicolumn{1}{l}{\textbf{CSCE-Net}} & ResNet50 & 0.869 & 0.887 & 0.897 & 0.807 & 0.852 & 0.888 & 0.794 & 0.827 & 0.839 \\
        \multicolumn{1}{l}{\textbf{CSCE-Net} } & VGG16 & 0.729 & 0.764 & 0.800 & 0.751 & 0.803 & 0.832 & 0.806 & 0.823 & 0.839 \\
        \multicolumn{1}{l}{\textbf{CSCE-Net} } & GCViT-XXT & 0.896 & 0.910 & 0.911 & 0.844 & 0.902 & 0.919 & 0.839 & 0.865 & 0.878  \\
\cline{1-11}
\end{tabular}

\label{t-1}
\end{table}

The mean average precision (mAP) results comparing with state-of-the-art methods are shown in Table \ref{t-1}. The CNNH \cite{xia2014supervised}, DNNH \cite{lai2015simultaneous}, DHN  \cite{zhu2016deep}, HashNet \cite{cao2017hashnet}, CSQ \cite{yuan2020central} and our proposed CSCE-Net are based on ResNet50 backbone; TransHash \cite{chen2021transhash}, CIBHash \cite{qiu2021CIBHash} use the Vision Transformer (ViT) as backbone.
From Table \ref{t-1}, it can be observed that our proposed CSCE-Net substantially outperforms all ResNet50 based comparison methods even better than transformer-based TransHash by up 8.4\% on ImageNet(16bits) and 14.4\% on NUS-WIDE(64bit). 
Specifically, compared with the state-of-the-art ResNet50 based methods CSQ, our CSCE-Net achieves average absolute boosts of 1.5\% in mAP for different bits on three benchmarks except for NUS-WIDE(16bit).
%Besides, with Res50 backbone, the state-of-the-art deep hashing methods CSQ, our CSCE-Net brings an increase of at least 
Especially, the performance boost on ImageNet100 dataset is much larger than that on others, which is very impressive. Note that ImageNet100 has the most categories with variations among three datasets. Therefore, our CSCE-Net has the strong capacity for extracting critical local detailed information guided by semantic global context.

On the other hand, our ResNet50 based comparisons are conducted on both single-label and multi-label datasets. As shown in results for multi-label MS COCO and NUS-WIDE, the performance boost of the proposed CSCE-Net is as obvious as that on ImageNet100 except for NUS-WIDE(16bit). We achieve 1.1\%, 1.4\% and 2.7\% higher mAP than the best competitor CSQ at 16-, 32- and 64-bit.
In summary, our CSCE-Net performs well in most cases of different bits, which confirms the representation capability of binary code for image retrieval.

\begin{figure}[htbp]
\centering
\includegraphics[width=0.8\textwidth]{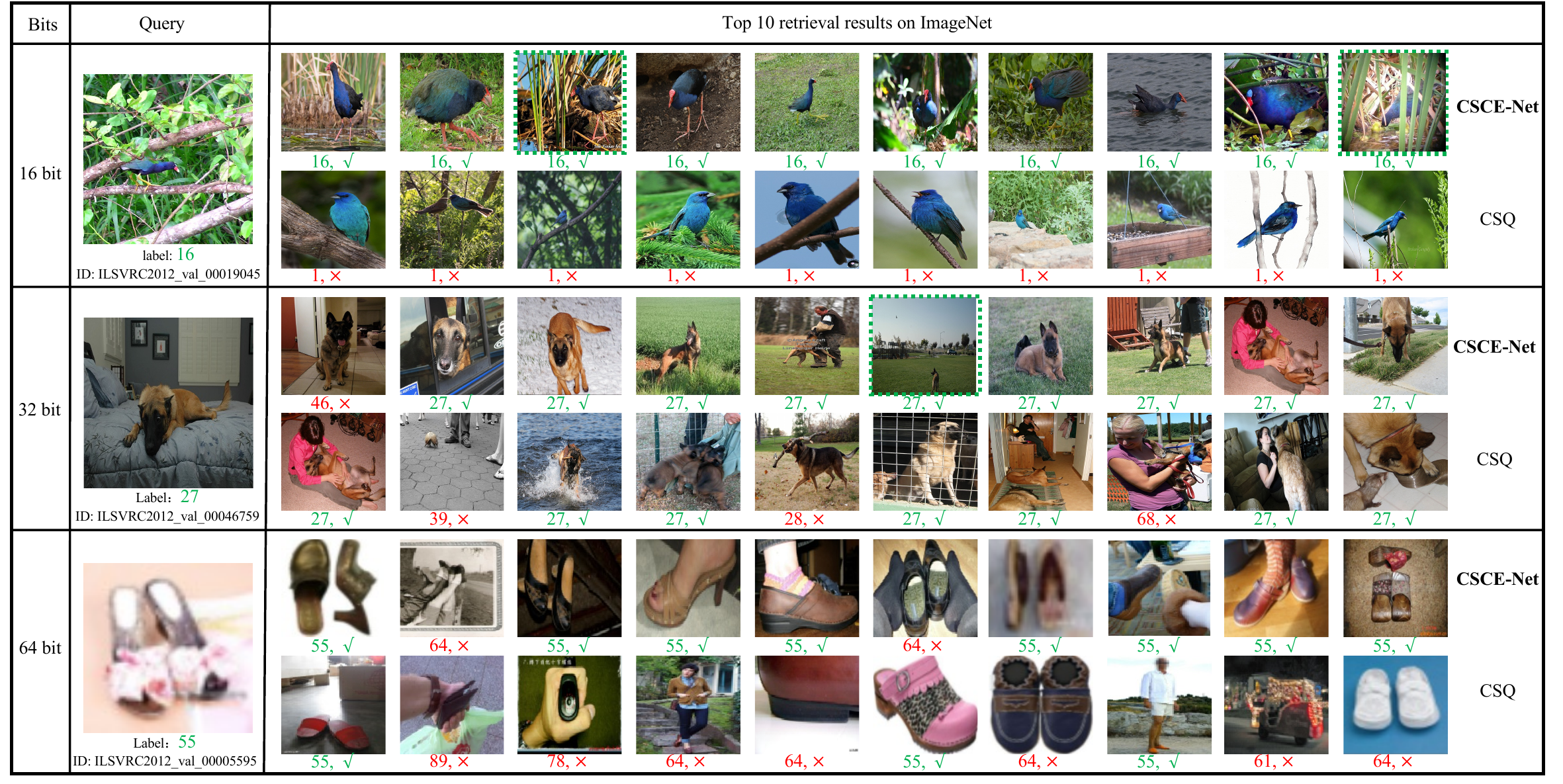}
\caption{Examples of top 10 retrieved images for three bits on Imagenet-100 dataset. Below each image, we mark the corresponding label in  \textcolor{green}{green} for accurate recall and \textcolor{red}{red} for wrong recall.}
\label{fig:fig4}
\end{figure}

\begin{figure}[ht]
\centering
\includegraphics[width=0.7\textwidth]{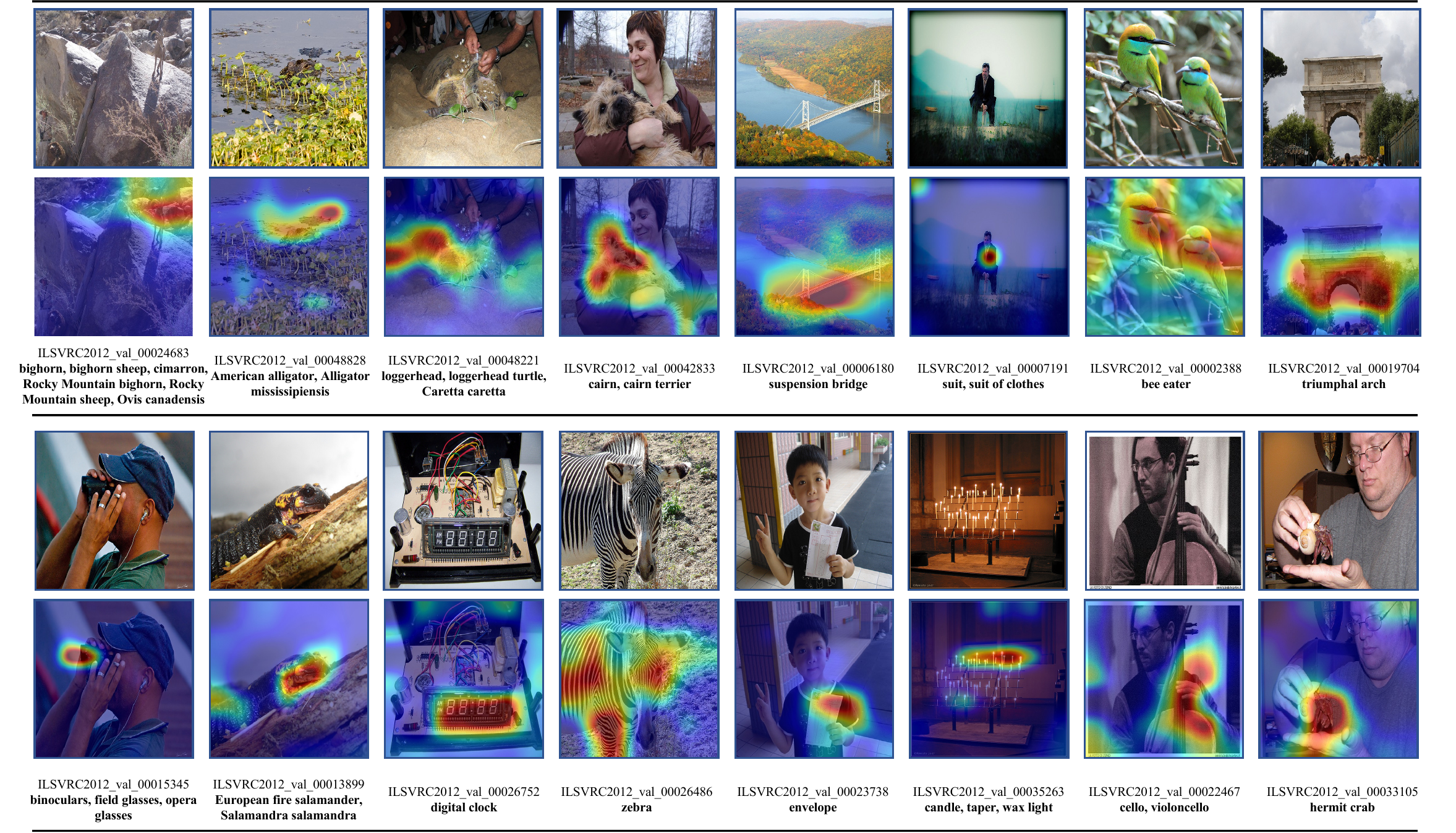}
\caption{Visualization of class activation map of the last non-reduced convolutional layer in CSCE-Net.
The image id and  bolded \textbf{label} are at the bottom of each image.}
\label{feamap}
\end{figure}

%We show the retrieval results on ImageNet in Fig. \ref{fig:fig4}. It can be seen that state-of- the-art methods CSQ with global feature will result in many false positives which are semantically similar to the query. Especially, in the first example, 

%It can be seen that proposed CSCE-Net can return much more relevant results. 
As shown in Fig.~\ref{fig:fig4}, each query has recalled a list of relative images, and our results are named by CSCE-Net. From the qualitative comparison, it is observed that the popular method CSQ and our CSCE-Net both can get accurate recalls in most cases. However, when the query image has a complex background or small main target like the first example of Fig.~\ref{fig:fig4}, our method performs much better than the other. Especially, the recall results of CSCE-Net also have redundant noise marked with green dashed boxes, which shows our method can deal with hard samples of unapparent targets in complex backgrounds. To validate this point, we visualize the last non-reduced convolutional
layer of CSCE-Net using TorchCAM \cite{torcham2020} shown in Fig.~\ref{feamap}.

\subsubsection{Ablation Studies}

\textbf{Analysis on the Effects of AIE.}
We conduct ablation study experiments to analyze which part of the network benefits our proposed method mainly. As shown in Table \ref{t-2}, using the same loss as mentioned in Section \ref{section_loss}, the baseline model is constructed on the 4-th layer of ResNet50, which represents the global context. Besides, we consume that the 3-rd layer feature namely ``local" has more detailed information compared to the 4-th layer with a much bigger feature map. In terms of that, the proposed AIE aligns global semantics and local details to suppress redundant context, which evaluates the performance on the final row in Table \ref{t-2}. The second row of ``Baseline + local" means the concatenation of local and global features shown as $f_{local}$ and $f_{global}$ in Fig.~\ref{fig:pipeline} to output continuous features. Compared with baseline result, it has an average 0.3\% absolute improvement benefiting from local details.

In addition, neither using a single sliding window nor concating sliding window features with different scales as shown in the 3rd-5th rows of Table \ref{t-2} has improved effectively. 
Compared with all conditions,  Our AIE module achieves relative boosts of 3\%, 1\% , 1\% in average mAP for each bit code on ImageNet100 and 7\%, 3\%, 2\% on MS COCO dataset, which confirms that the detailed feature from local branch is aligned and filtered by global semantic context to focus on important instances area. Especially for 16 bit code, the proposed AIE module strongly improves performance with fewer hash values. Therefore, in the case where only less information can be retained, our model still has a strong representation of critical content.

\begin{table}
\centering
\scriptsize
\caption{Experimental results of CSCE-Net with variants structure.}
\begin{tabular}{cc|ccc|ccc}
    \hline
    \multicolumn{2}{l}{\multirow{2}*{Struct}} & \multicolumn{3}{c|}{ImageNet(mAP@1000)}  & \multicolumn{3}{c}{MS COCO(mAP@5000)}\\
    \cline{3-8} 
    \multicolumn{2}{c}{} & 16bits & 32bits & 64bits &16 bits & 32bits & 64bits \\
    \toprule
        \multicolumn{2}{l}{Baseline} & 0.839 & 0.879 & 0.889  & 0.752 & 0.829 & 0.865\\
        \multicolumn{2}{l}{Baseline + local} & 0.840 & 0.883  & 0.894   & 0.760 & 0.828 & 0.869\\
        \multicolumn{2}{l}{Baseline + global-sw2} & 0.856 & 0.868 & 0.888  & 0.739 & 0.802 & 0.851\\
        \multicolumn{2}{l}{Baseline + global-sw4} & 0.857 & 0.870 & 0.882  & 0.754 & 0.812 & 0.841\\
        \multicolumn{2}{l}{Baseline + global-sw2 + global-sw4} & 0.854 & 0.879 & 0.894  & 0.768 & 0.814 & 0.854\\
        \multicolumn{2}{l}{Baseline + AIE} & \textbf{0.869} & \textbf{0.887} & \textbf{0.897}  & \textbf{0.807} & \textbf{0.852} & \textbf{0.888}\\
\hline
\end{tabular}
\label{t-2}
\end{table}

\begin{table}
\scriptsize
\centering
\caption {mAP comparison with variants loss of our method.}
\begin{tabular}{cc|ccc}
    \hline
    \multicolumn{2}{l}{\multirow{2}*{Loss}} & \multicolumn{3}{c}{Imagenet100(mAP@1000)} \\
    \cline{3-5} 
    \multicolumn{2}{c}{} & 16 bits & 32bits & 64bits  \\
    \toprule  % 			
        \multicolumn{2}{l}{CE} & 0.845 & 0.877 & 0.887  \\
        \multicolumn{2}{l}{CE+Qua} & 0.860 & 0.885 & 0.889  \\
        \multicolumn{2}{l}{CF} & 0.860 & 0.877 & 0.891  \\
     %   \multicolumn{2}{c}{CE+Center} & 0. & 0.871 & 0.872 & 0.353 & 0.522 & 0.610 \\
        \multicolumn{2}{l}{CF+Qua} & 0.862 & 0.884 & 0.894\\
        \multicolumn{2}{l}{CF+DSF} & \textbf{0.869} & \textbf{0.887} & \textbf{0.897} \\
        
\hline
\end{tabular}
\label{t-3}
\end{table}

\textbf{Analysis on the Effects of DSF and loss functions.}
We have explained that the target of hash encoding is to cluster all samples in the same category. Besides, it is desirable that the output hash code is balanced with 0 and 1 as much as possible. Therefore, several experiments with different loss settings were conducted as shown in Table \ref{t-3}. ``CE" means using a cross-entropy function to measure the loss between continuous features and ground truth binary codes. ``CF" and ``Qua" means CosFace loss and Quantization loss with the simple $sign$ function we mentioned at Section \ref{section_loss} respectively. DSF is the proposed learnable sign function for quantization loss. From Table \ref{t-3}, we can see that the proposed CF+DSF algorithm achieves the best performance on the Imagenet100 dataset with the average mAP of 0.869, 0.887 and 0.897 respectively. As mentioned previously, the softmax cross-entropy loss performs well on classification, but is not optimal for feature learning due to its large intra-class variations. In contrast, the proposed method using cosine similarity to measure code distance is more adaptive. 
Besides, the comparisons of ``no quantization loss", ``simple quantization loss" and ``DSF quantization loss" shown in the last 3 rows of Table \ref{t-3} draw a conclusion that it is necessary for reducing information loss through sign function. As for mAP results, our learnable sign function has a stable improvement.

\section{Conclusion}
In this paper, we consider the representation capability of binary code with a novel feature extraction module for deep hashing. The proposed Cross-Scale Context Extracted Hashing Network (CSCE-Net) can improve the quality of handling unapparent targets and complex backgrounds by utilizing global semantics to extract fine-grained local information. Moreover, we leverage the concept of code balance to inherit the stability from continuous float features to retrieval binary codes through a learnable dynamic sign function.
Extensive experiments validated the efficiency of our method in both single-label and multi-label retrieval benchmarks. 
As part of the future work, we are exploring how to learn better feature representations for image retrieval in an unsupervised way.

% \section{Section Title}
% Main contents here.

% \subsection{Subsection Title}
% A figure in Fig.~\ref{fig:spiral}. Please use high quality graphics for your camera-ready submission -- if you can use a vector graphics format such as \texttt{.eps} or \texttt{.pdf}.
% \begin{figure}[htp]
% \begin{center}
% \includegraphics[width=0.5\textwidth]{spiral.eps}
% \caption{A spiral.}\label{fig:spiral}
% \end{center}
% \end{figure}

% An example of citation~\cite{DBLP:conf/acml/2009}.

% \acks{Acknowledgements should go at the end, before appendices and references.}

%\bibliographystyle{plain}
% \bibliography{acml21}
\bibliography{sample}

\begin{thebibliography}{40}
\providecommand{\natexlab}[1]{#1}
\providecommand{\url}[1]{\texttt{#1}}
\expandafter\ifx\csname urlstyle\endcsname\relax
  \providecommand{\doi}[1]{doi: #1}\else
  \providecommand{\doi}{doi: \begingroup \urlstyle{rm}\Url}\fi

\bibitem[Balntas et~al.(2016)Balntas, Riba, Ponsa, and
  Mikolajczyk]{balntas2016learning}
Vassileios Balntas, Edgar Riba, Daniel Ponsa, and Krystian Mikolajczyk.
\newblock Learning local feature descriptors with triplets and shallow
  convolutional neural networks.
\newblock In \emph{Bmvc}, volume~1, page~3, 2016.

\bibitem[Cao et~al.(2020)Cao, Araujo, and Sim]{cao2020unifying}
Bingyi Cao, Andre Araujo, and Jack Sim.
\newblock Unifying deep local and global features for image search.
\newblock In \emph{European Conference on Computer Vision}, pages 726--743.
  Springer, 2020.

\bibitem[Cao et~al.(2018)Cao, Long, Liu, and Wang]{cao2018deep}
Yue Cao, Mingsheng Long, Bin Liu, and Jianmin Wang.
\newblock Deep cauchy hashing for hamming space retrieval.
\newblock In \emph{Proceedings of the IEEE Conference on Computer Vision and
  Pattern Recognition}, pages 1229--1237, 2018.

\bibitem[Cao et~al.(2017)Cao, Long, Wang, and Yu]{cao2017hashnet}
Zhangjie Cao, Mingsheng Long, Jianmin Wang, and Philip~S Yu.
\newblock Hashnet: Deep learning to hash by continuation.
\newblock In \emph{Proceedings of the IEEE international conference on computer
  vision}, pages 5608--5617, 2017.

\bibitem[Chen et~al.(2017)Chen, Papandreou, Schroff, and Adam]{chen2017aspp}
Liang-Chieh Chen, George Papandreou, Florian Schroff, and Hartwig Adam.
\newblock Rethinking atrous convolution for semantic image segmentation.
\newblock \emph{arXiv preprint arXiv:1706.05587}, 2017.

\bibitem[Chen et~al.(2021)Chen, Zhang, Liu, Chang, Ye, and
  Qi]{chen2021transhash}
Yongbiao Chen, Sheng Zhang, Fangxin Liu, Zhigang Chang, Mang Ye, and Zhengwei
  Qi.
\newblock Transhash: Transformer-based hamming hashing for efficient image
  retrieval.
\newblock \emph{arXiv preprint arXiv:2105.01823}, 2021.

\bibitem[Chua et~al.(2009)Chua, Tang, Hong, Li, Luo, and Zheng]{chua2009nus}
Tat-Seng Chua, Jinhui Tang, Richang Hong, Haojie Li, Zhiping Luo, and Yantao
  Zheng.
\newblock Nus-wide: a real-world web image database from national university of
  singapore.
\newblock In \emph{Proceedings of the ACM international conference on image and
  video retrieval}, pages 1--9, 2009.

\bibitem[Deng et~al.(2019)Deng, Guo, Xue, and Zafeiriou]{deng2019arcface}
Jiankang Deng, Jia Guo, Niannan Xue, and Stefanos Zafeiriou.
\newblock Arcface: Additive angular margin loss for deep face recognition.
\newblock In \emph{Proceedings of the IEEE/CVF conference on computer vision
  and pattern recognition}, pages 4690--4699, 2019.

\bibitem[Fernandez(2020)]{torcham2020}
François-Guillaume Fernandez.
\newblock Torchcam: class activation explorer.
\newblock \url{https://github.com/frgfm/torch-cam}, March 2020.

\bibitem[He et~al.(2016)He, Zhang, Ren, and Sun]{he2016resnet}
Kaiming He, Xiangyu Zhang, Shaoqing Ren, and Jian Sun.
\newblock Deep residual learning for image recognition.
\newblock In \emph{Proceedings of the IEEE conference on computer vision and
  pattern recognition}, pages 770--778, 2016.

\bibitem[Hoe et~al.(2021)Hoe, Ng, Zhang, Chan, Song, and Xiang]{hoe2021one}
Jiun~Tian Hoe, Kam~Woh Ng, Tianyu Zhang, Chee~Seng Chan, Yi-Zhe Song, and Tao
  Xiang.
\newblock One loss for all: Deep hashing with a single cosine similarity based
  learning objective.
\newblock \emph{Advances in Neural Information Processing Systems}, 34, 2021.

\bibitem[Krizhevsky et~al.(2012)Krizhevsky, Sutskever, and
  Hinton]{krizhevsky2012imagenet}
Alex Krizhevsky, Ilya Sutskever, and Geoffrey~E Hinton.
\newblock Imagenet classification with deep convolutional neural networks.
\newblock \emph{Advances in neural information processing systems}, 25, 2012.

\bibitem[Lai et~al.(2015)Lai, Pan, Liu, and Yan]{lai2015simultaneous}
Hanjiang Lai, Yan Pan, Ye~Liu, and Shuicheng Yan.
\newblock Simultaneous feature learning and hash coding with deep neural
  networks.
\newblock In \emph{Proceedings of the IEEE conference on computer vision and
  pattern recognition}, pages 3270--3278, 2015.

\bibitem[Li et~al.(2017)Li, Sun, He, and Tan]{li2017deep}
Qi~Li, Zhenan Sun, Ran He, and Tieniu Tan.
\newblock Deep supervised discrete hashing.
\newblock \emph{Advances in neural information processing systems}, 30, 2017.

\bibitem[Li et~al.(2015)Li, Wang, and Kang]{li2015feature}
Wu-Jun Li, Sheng Wang, and Wang-Cheng Kang.
\newblock Feature learning based deep supervised hashing with pairwise labels.
\newblock \emph{arXiv preprint arXiv:1511.03855}, 2015.

\bibitem[Lin et~al.(2014)Lin, Maire, Belongie, Hays, Perona, Ramanan,
  Doll{\'a}r, and Zitnick]{lin2014microsoft}
Tsung-Yi Lin, Michael Maire, Serge Belongie, James Hays, Pietro Perona, Deva
  Ramanan, Piotr Doll{\'a}r, and C~Lawrence Zitnick.
\newblock Microsoft coco: Common objects in context.
\newblock In \emph{European conference on computer vision}, pages 740--755.
  Springer, 2014.

\bibitem[Liong et~al.(2016)Liong, Lu, Tan, and Zhou]{liong2016deep}
Venice~Erin Liong, Jiwen Lu, Yap-Peng Tan, and Jie Zhou.
\newblock Deep video hashing.
\newblock \emph{IEEE Transactions on Multimedia}, 19\penalty0 (6):\penalty0
  1209--1219, 2016.

\bibitem[Liu et~al.(2016)Liu, Wang, Shan, and Chen]{liu2016deep}
Haomiao Liu, Ruiping Wang, Shiguang Shan, and Xilin Chen.
\newblock Deep supervised hashing for fast image retrieval.
\newblock In \emph{Proceedings of the IEEE conference on computer vision and
  pattern recognition}, pages 2064--2072, 2016.

\bibitem[Liu et~al.(2017)Liu, Wen, Yu, Li, Raj, and Song]{liu2017sphereface}
Weiyang Liu, Yandong Wen, Zhiding Yu, Ming Li, Bhiksha Raj, and Le~Song.
\newblock Sphereface: Deep hypersphere embedding for face recognition.
\newblock In \emph{Proceedings of the IEEE conference on computer vision and
  pattern recognition}, pages 212--220, 2017.

\bibitem[Noh et~al.(2017{\natexlab{a}})Noh, Araujo, Sim, Weyand, and
  Han]{noh2017large}
Hyeonwoo Noh, Andre Araujo, Jack Sim, Tobias Weyand, and Bohyung Han.
\newblock Large-scale image retrieval with attentive deep local features.
\newblock In \emph{Proceedings of the IEEE international conference on computer
  vision}, pages 3456--3465, 2017{\natexlab{a}}.

\bibitem[Noh et~al.(2017{\natexlab{b}})Noh, Araujo, Sim, Weyand, and
  Han]{noh2017selfatt}
Hyeonwoo Noh, Andre Araujo, Jack Sim, Tobias Weyand, and Bohyung Han.
\newblock Large-scale image retrieval with attentive deep local features.
\newblock In \emph{Proceedings of the IEEE international conference on computer
  vision}, pages 3456--3465, 2017{\natexlab{b}}.

\bibitem[Qiu et~al.(2021)Qiu, Su, Ou, Yu, and Chen]{qiu2021CIBHash}
Zexuan Qiu, Qinliang Su, Zijing Ou, Jianxing Yu, and Changyou Chen.
\newblock Unsupervised hashing with contrastive information bottleneck.
\newblock \emph{arXiv preprint arXiv:2105.06138}, 2021.

\bibitem[Revaud et~al.(2019)Revaud, Weinzaepfel, De~Souza, Pion, Csurka, Cabon,
  and Humenberger]{revaud2019r2d2}
Jerome Revaud, Philippe Weinzaepfel, C{\'e}sar De~Souza, Noe Pion, Gabriela
  Csurka, Yohann Cabon, and Martin Humenberger.
\newblock R2d2: repeatable and reliable detector and descriptor.
\newblock \emph{arXiv preprint arXiv:1906.06195}, 2019.

\bibitem[Rumelhart et~al.(1986)Rumelhart, Hinton, and
  Williams]{rumelhart1986learning}
David~E Rumelhart, Geoffrey~E Hinton, and Ronald~J Williams.
\newblock Learning representations by back-propagating errors.
\newblock \emph{nature}, 323\penalty0 (6088):\penalty0 533--536, 1986.

\bibitem[Russakovsky et~al.(2015)Russakovsky, Deng, Su, Krause, Satheesh, Ma,
  Huang, Karpathy, Khosla, Bernstein, et~al.]{russakovsky2015imagenet}
Olga Russakovsky, Jia Deng, Hao Su, Jonathan Krause, Sanjeev Satheesh, Sean Ma,
  Zhiheng Huang, Andrej Karpathy, Aditya Khosla, Michael Bernstein, et~al.
\newblock Imagenet large scale visual recognition challenge.
\newblock \emph{International journal of computer vision}, 115\penalty0
  (3):\penalty0 211--252, 2015.

\bibitem[Shen et~al.(2015)Shen, Shen, Liu, and Shen]{2015Supervised}
F.~Shen, C.~Shen, W.~Liu, and H.~T. Shen.
\newblock Supervised discrete hashing.
\newblock \emph{IEEE}, 2015.

\bibitem[Sim{\'e}oni et~al.(2019)Sim{\'e}oni, Avrithis, and
  Chum]{simeoni2019local}
Oriane Sim{\'e}oni, Yannis Avrithis, and Ondrej Chum.
\newblock Local features and visual words emerge in activations.
\newblock In \emph{Proceedings of the IEEE/CVF Conference on Computer Vision
  and Pattern Recognition}, pages 11651--11660, 2019.

\bibitem[Sylvester(1867)]{sylvester1867lx}
James~Joseph Sylvester.
\newblock Lx. thoughts on inverse orthogonal matrices, simultaneous
  signsuccessions, and tessellated pavements in two or more colours, with
  applications to newton's rule, ornamental tile-work, and the theory of
  numbers.
\newblock \emph{The London, Edinburgh, and Dublin Philosophical Magazine and
  Journal of Science}, 34\penalty0 (232):\penalty0 461--475, 1867.

\bibitem[Vaswani et~al.(2017)Vaswani, Shazeer, Parmar, Uszkoreit, Jones, Gomez,
  Kaiser, and Polosukhin]{vaswani2017attention}
Ashish Vaswani, Noam Shazeer, Niki Parmar, Jakob Uszkoreit, Llion Jones,
  Aidan~N Gomez, {\L}ukasz Kaiser, and Illia Polosukhin.
\newblock Attention is all you need.
\newblock \emph{Advances in neural information processing systems}, 30, 2017.

\bibitem[Wang et~al.(2018)Wang, Wang, Zhou, Ji, Gong, Zhou, Li, and
  Liu]{wang2018cosface}
Hao Wang, Yitong Wang, Zheng Zhou, Xing Ji, Dihong Gong, Jingchao Zhou, Zhifeng
  Li, and Wei Liu.
\newblock Cosface: Large margin cosine loss for deep face recognition.
\newblock In \emph{Proceedings of the IEEE conference on computer vision and
  pattern recognition}, pages 5265--5274, 2018.

\bibitem[Wang et~al.(2021)Wang, Chen, Zhang, Meng, Liang, and
  Xia]{wang2021weakly}
Jinpeng Wang, Bin Chen, Qiang Zhang, Zaiqiao Meng, Shangsong Liang, and Shutao
  Xia.
\newblock Weakly supervised deep hyperspherical quantization for image
  retrieval.
\newblock In \emph{Proceedings of the AAAI Conference on Artificial
  Intelligence}, volume~35, pages 2755--2763, 2021.

\bibitem[Wang et~al.(2015)Wang, Liu, Kumar, and Chang]{wang2015learning}
Jun Wang, Wei Liu, Sanjiv Kumar, and Shih-Fu Chang.
\newblock Learning to hash for indexing big data—a survey.
\newblock \emph{Proceedings of the IEEE}, 104\penalty0 (1):\penalty0 34--57,
  2015.

\bibitem[Wieczorek et~al.(2021)Wieczorek, Rychalska, and
  D{a}browski]{wieczorek2021unreasonable}
Miko{\l}aj Wieczorek, Barbara Rychalska, and Jacek D{a}browski.
\newblock On the unreasonable effectiveness of centroids in image retrieval.
\newblock In \emph{International Conference on Neural Information Processing},
  pages 212--223. Springer, 2021.

\bibitem[Xia et~al.(2014)Xia, Pan, Lai, Liu, and Yan]{xia2014supervised}
Rongkai Xia, Yan Pan, Hanjiang Lai, Cong Liu, and Shuicheng Yan.
\newblock Supervised hashing for image retrieval via image representation
  learning.
\newblock In \emph{Twenty-eighth AAAI conference on artificial intelligence},
  2014.

\bibitem[Xu et~al.(2021)Xu, Xu, Chai, Li, Zuo, Yang, and Yuan]{xu2021hhf}
Chengyin Xu, Zhengzhuo Xu, Zenghao Chai, Hongjia Li, Qiruyi Zuo, Lingyu Yang,
  and Chun Yuan.
\newblock Hhf: Hashing-guided hinge function for deep hashing retrieval.
\newblock \emph{arXiv preprint arXiv:2112.02225}, 2021.

\bibitem[Yang et~al.(2021)Yang, He, Fan, Shi, Xue, Li, Ding, and
  Huang]{yang2021dolg}
Min Yang, Dongliang He, Miao Fan, Baorong Shi, Xuetong Xue, Fu~Li, Errui Ding,
  and Jizhou Huang.
\newblock Dolg: Single-stage image retrieval with deep orthogonal fusion of
  local and global features.
\newblock In \emph{Proceedings of the IEEE/CVF International Conference on
  Computer Vision}, pages 11772--11781, 2021.

\bibitem[Yuan et~al.(2020)Yuan, Wang, Zhang, Tay, Jie, Liu, and
  Feng]{yuan2020central}
Li~Yuan, Tao Wang, Xiaopeng Zhang, Francis~EH Tay, Zequn Jie, Wei Liu, and
  Jiashi Feng.
\newblock Central similarity quantization for efficient image and video
  retrieval.
\newblock In \emph{Proceedings of the IEEE/CVF Conference on Computer Vision
  and Pattern Recognition}, pages 3083--3092, 2020.

\bibitem[Zhang et~al.(2021)Zhang, Peng, and Li]{zhang2021instance}
Zhiwei Zhang, Allen Peng, and Hongsheng Li.
\newblock Instance-weighted central similarity for multi-label image retrieval.
\newblock \emph{arXiv preprint arXiv:2108.05274}, 2021.

\bibitem[Zhe et~al.(2019)Zhe, Chen, and Yan]{zhe2019deep}
Xuefei Zhe, Shifeng Chen, and Hong Yan.
\newblock Deep class-wise hashing: Semantics-preserving hashing via class-wise
  loss.
\newblock \emph{IEEE transactions on neural networks and learning systems},
  31\penalty0 (5):\penalty0 1681--1695, 2019.

\bibitem[Zhu et~al.(2016)Zhu, Long, Wang, and Cao]{zhu2016deep}
Han Zhu, Mingsheng Long, Jianmin Wang, and Yue Cao.
\newblock Deep hashing network for efficient similarity retrieval.
\newblock In \emph{Proceedings of the AAAI conference on Artificial
  Intelligence}, volume~30, 2016.

\end{thebibliography}
\appendix

% \section{First Appendix}\label{apd:first}

% This is the first appendix.

% \section{Second Appendix}\label{apd:second}

% This is the second appendix.

\end{document}